%% file: main.tex
\documentclass[conference]{IEEEtran}
\usepackage{graphicx}
\usepackage{amsmath}
\usepackage[numbers,compress]{natbib}
\usepackage{hyperref}
\usepackage{caption}
\usepackage{array,longtable}
\usepackage{amssymb}
\usepackage{float}
\usepackage{booktabs}
\usepackage{cleveref}

\usepackage[dvipsnames]{xcolor}

\definecolor{light_green}{RGB}{206, 255, 217}
\definecolor{light_red}{RGB}{255, 203, 203}
\definecolor{light_gray}{RGB}{250, 250, 250}
\definecolor{light_orange}{RGB}{255, 241, 207}
\definecolor{light_blue}{RGB}{207, 233, 255}
\definecolor{light_purple}{RGB}{221, 207, 255}
\definecolor{light_pink}{RGB}{247, 215, 255}
\definecolor{light_blue_2}{RGB}{215, 249, 255}

\Crefname{figure}{Fig.}{Figs.}

\usepackage{tikz}
\usepackage{pgfplots}
\usepackage{pgf-pie}
\pgfplotsset{compat=1.18}

\title{Teaching Machine Learning Fundamentals with LEGO Robotics}

\author{
    \IEEEauthorblockN{Viacheslav Sydora, Guner Dilsad Er, Michael Muehlebach}
    \IEEEauthorblockA{Max Planck Institute for Intelligent Systems, T\"ubingen, Germany \\
    \{viacheslav.sydora, guenerdilsad.er, michael.muehlebach\}@tuebingen.mpg.de}
}

\begin{document}
\maketitle

\begin{abstract}
This paper presents the web-based platform \textit{Machine Learning with Bricks} and an accompanying two-day course designed to teach machine learning concepts to students aged 12 to 17 through programming-free robotics activities. \textit{Machine Learning with Bricks} is an open source platform and combines interactive visualizations with LEGO robotics to teach three core algorithms: KNN, linear regression, and Q-learning. Students learn by collecting data, training models, and interacting with robots via a web-based interface. Pre- and post-surveys with 14 students indicate statistically significant improvements in self-reported understanding of machine learning algorithms, changes in AI-related terminology toward more technical language, high platform usability, and increased motivation for continued learning. This work suggests that tangible, visualization-based approaches can make machine learning concepts accessible and engaging for young learners while maintaining technical depth. The platform is freely available at \url{https://learning-and-dynamics.github.io/ml-with-bricks/}, with video tutorials guiding students through the experiments at \url{https://youtube.com/playlist?list=PLx1grFu4zAcwfKKJZ1Ux4LwRqaePCOA2J}.

\end{abstract}

\section*{Supplementary Materials}

The survey instruments, anonymized response data, and analysis code used to evaluate the course and platform are publicly available at \url{https://github.com/learning-and-dynamics/ml-with-bricks/tree/main/survey}.

\section{Introduction}
AI education is crucial for enabling K-12 students to understand, use, and develop AI systems \cite{heintz2021}. This imperative has sparked research into effective pedagogical approaches and prompted countries worldwide to develop AI curricula \cite{casal-otero2023}.\looseness=-1

Most existing work focuses on ethics or AI applications, while underlying algorithms remain inaccessible due to their complexity.
These technical aspects are typically introduced only at the university level, with AI systems treated as `black boxes' in K-12 education \cite{sanusi2023}. Yet, research with adult users has shown that revealing the internal mechanisms of machine learning models can significantly improve understanding \cite{kulesza2015}. Moreover, while most courses use block-based programming rather than classical programming \cite{kahn2018, van_brummelen2021}, this still introduces cognitive load and shifts students' attention toward implementation details rather than the underlying concepts.

To address these challenges, researchers have explored visualizations \cite{evangelista2018, broll2022, reddy2021, wan2020} and tangible robotics \cite{zhang2023, touretzky2018, williams2019, lin2020, absalon2025} as ways to make AI concepts more accessible to younger learners. However, most existing platforms either focus on a single algorithm \cite{wan2020, zhang2023} or depend on custom robotics systems that are not widely available in schools \cite{touretzky2018, williams2019, lin2020}.

We address this gap by developing a web-based platform \textit{Machine Learning with Bricks} that integrates interactive data and algorithm visualizations with LEGO robotics, a widely available tool in schools that requires no additional purchase, to teach algorithms such as KNN, linear regression, and Q-learning to students aged 12 to 17. Using this platform, we conducted a two-day workshop with children to evaluate the effectiveness of our approach. Our key contributions are as~follows:
\begin{enumerate}
    \item[C1.] \textit{Machine Learning with Bricks}\footnote{\url{https://learning-and-dynamics.github.io/ml-with-bricks/}}: an open-source web-based platform that combines interactive visualizations and video demonstrations of experiments with LEGO robotics to teach core machine learning algorithms without programming.\looseness=-1
    \item[C2.] A two-day workshop curriculum featuring three hands-on activities that teach KNN, linear regression, and Q-learning through tangible robot interaction, structured educational scaffolding, and real-time data and algorithm visualization.\looseness=-1
    \item[C3.] Empirical evaluation indicating statistically significant improvements in self-reported understanding of machine learning concepts, shifts in AI-related terminology toward more technical language, high platform usability, and increased motivation for continued learning.
\end{enumerate}

\section{Related Work}
\label{sec:related_work}

Given the importance of early AI introduction in helping children understand the modern world around them \cite{hitron2018}, AI literacy is becoming as essential as writing skills \cite{kandlhofer2016}. Consequently, foundational frameworks have been established to guide AI education. These include five core AI concepts designed for universal childhood understanding \cite{touretzky2019} and a comprehensive set of K-12 AI literacy competencies that incorporate specific design considerations for educational technologies \cite{long2020}. Building on these frameworks, numerous course proposals have emerged \cite{sanusi2023}. For instance, \citet{van_brummelen2021} build a course specifically to teach high-schoolers eight competencies \cite{long2020} by having students develop conversational agents with the MIT App Inventor, noting that machine learning and ethics competencies proved most challenging for students to grasp according to the survey results. 

The complexity of machine learning concepts requires learning artifacts that support scaffolding students' conceptual understanding \cite{ng2021}. Given that visualizations have proven effective in other disciplines \cite{wieman2008}, researchers have explored visual tools for this purpose. These tools make machine learning more accessible by demystifying fundamental concepts such as clustering \cite{evangelista2018} and knowledge representation \cite{lin2020}, as well as specific algorithms including k-means \cite{wan2020}, gradient descent \cite{broll2022}, KNN \cite{reddy2021, absalon2025}, and Q-learning \cite{zhang2023}. For instance, SmileyCluster \cite{wan2020} is a web-based collaborative learning environment that utilizes glyph-based data visualization and superposition to help K-12 students understand k-means clustering through a face-overlay metaphor that translates multi-dimensional data attributes into facial features. 

To make machine learning more accessible to younger audiences, several studies have applied Papert's constructionism \cite{papert1991, papert1980} by engaging students in hands-on data collection using tangible sensors for gesture recognition \cite{zimmermann-niefield2020, hitron2019}, athletic movement classification \cite{zimmermann-niefield2019}, and drawing recognition \cite{kaspersen2021}. Beyond data collection, other researchers have developed interactive tangible devices that allow children to engage directly with machine learning platforms throughout the learning process \cite{williams2019, lin2020, williams2021, burgsteiner2016}.\looseness=-1

Some studies have also utilized widespread LEGO robotics platforms to teach machine learning concepts \cite{zhang2023, absalon2025, parsons2004}. For example, \citet{zhang2023} proposes a platform that combines LEGO SPIKE Prime robots with mobile augmented reality interfaces to introduce reinforcement learning to K-12 students through a treasure hunting activity, allowing users to visualize Q-tables, state-action pairs, and the robot's learning process in real-time overlaid on the physical environment. Most recently, AlphAI \cite{absalon2025} was developed to make machine learning algorithms accessible by combining hands-on training of educational robots with real-time visualizations of algorithm details such as neural networks, KNN, and Q-learning.

Building on these foundations, our work synthesizes effective pedagogical strategies from prior research by combining student-led data collection, interactive visualizations, and tangible robotics into a unified, programming-free platform that utilizes off-the-shelf LEGO kits to make machine learning fundamentals fully transparent and intuitively accessible to young learners.

\section{Learning Platform}
\label{sec:learning_platform}

We developed \textit{Machine Learning with Bricks}, a web-based platform that enables hands-on machine learning experiments using LEGO Education SPIKE Prime kits. The platform is designed for both individual use at home and instruction in the classroom. It establishes wireless connectivity with the robotic kit, enabling real-time visualization of student-collected data.

The platform provides interactive implementations of three basic machine learning algorithms: KNN, linear regression, and Q-learning. These were selected such that their underlying mathematical principles, such as coordinate systems, line equations, and tabular value updates, are accessible to students, allowing learners to develop mechanistic understanding rather than treating models as black boxes. Furthermore, the three algorithms represent distinct machine learning paradigms, providing broad coverage of foundational concepts.

This section introduces the platform design, followed by detailed descriptions of three robots designed to illustrate the algorithms: the \textit{Fruit Detector} for KNN, the \textit{Pitcher} for linear regression, and the \textit{Crawler} for Q-learning.

\subsection{Platform design}
\begin{figure}[!ht]
    \centering
    \includegraphics[width=\columnwidth]{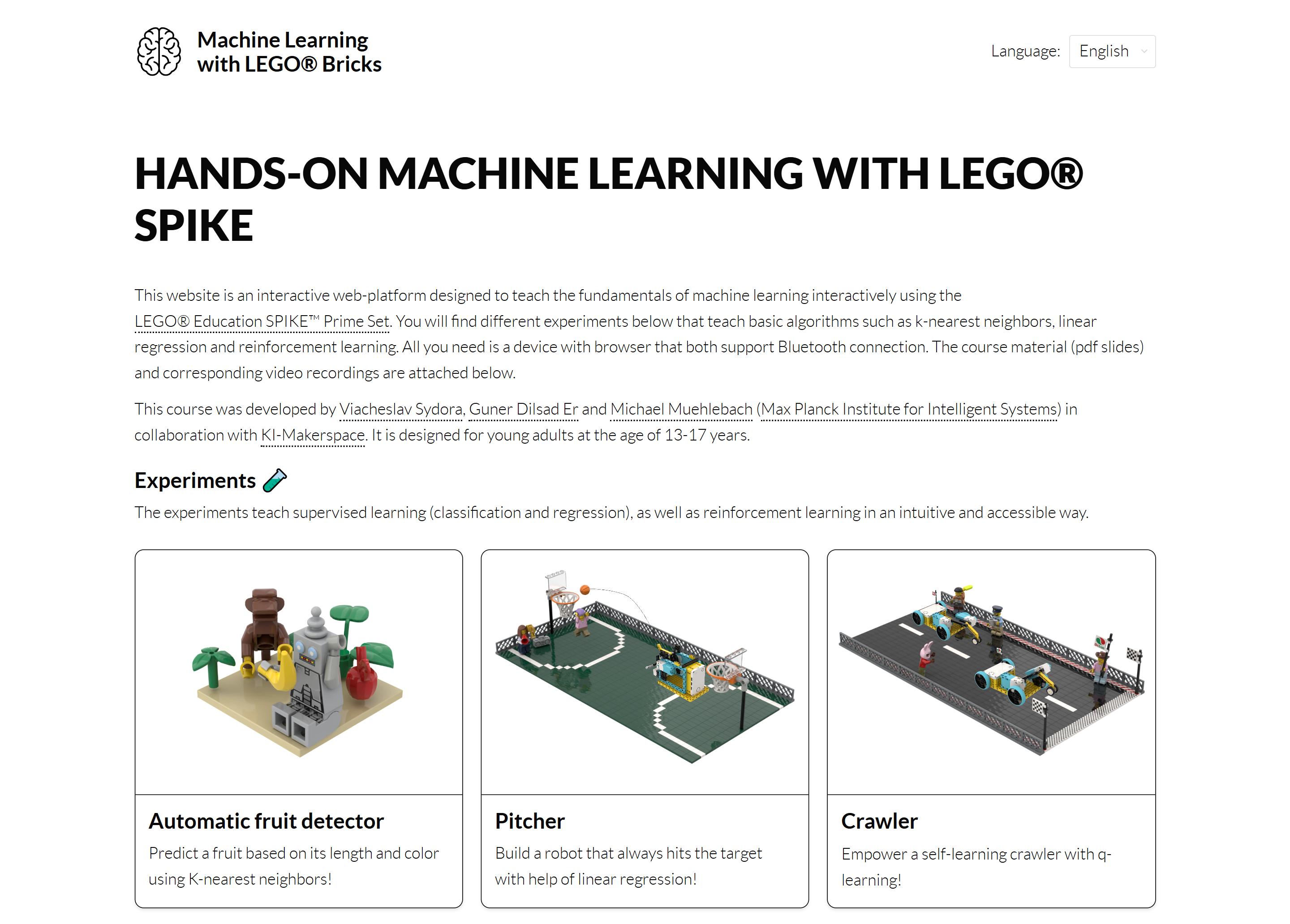}
    \caption{Landing page of the \textit{Machine Learning with Bricks} platform showing the available experiments.}
    \label{fig:platform_screenshot}
\end{figure}

The platform is an open-source web-based educational tool available in English and German under \url{https://learning-and-dynamics.github.io/ml-with-bricks/}. The website features a kid-friendly user interface as shown in \Cref{fig:platform_screenshot} and includes a landing page, a connection troubleshooting page for LEGO kits, and individual experiment pages, each dedicated to a specific learning activity.

Each experiment page provides step-by-step instructions on conducting the experiment, including guidance for robot assembly, connecting the LEGO hub to the platform, performing data collection, model training, and inference. The instructional materials are complemented by video tutorials\footnote{\href{https://youtube.com/playlist?list=PLx1grFu4zAcwfKKJZ1Ux4LwRqaePCOA2J}{https://youtube.com/playlist?list=PLx1grFu4zAcwfKKJZ1Ux4LwRqaePC\\OA2J}}, which visually guide learners through the experimental workflow. Each page features an interactive block comprising visualizations that display the data collected by the students and illustrate the algorithm in action, as well as a connection block responsible for communicating with the LEGO hub via Bluetooth. In addition to the simple pairing process, the experiment pages are robust to reloads and hub disconnections, ensuring accessibility for younger learners.

\subsection{Fruit detector} 

\begin{figure}[!ht]
    \centering
    \includegraphics[width=\columnwidth]{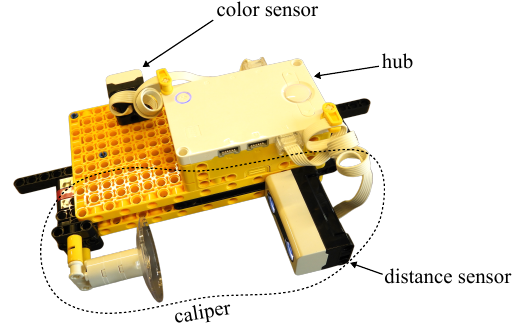}
    \caption{\textit{Fruit detector} consisting of a hub, a color sensor, and a caliper mechanism equipped with a distance sensor.}
    \label{fig:fruit_detector_photo}
\end{figure}

\textbf{Device design.} The \textit{Fruit Detector} is depicted in \Cref{fig:fruit_detector_photo}. It enables fruit detection of bananas and apples based on two features: color and length. Color is quantified using the green channel intensity of the RGB color model, ranging from 0 (red) to 255 (yellow), while length is measured with an ultrasonic sensor integrated into a caliper-like mechanism. Both sensors are connected through a central hub.

\begin{figure}[!ht]
    \centering
    \includegraphics[width=\columnwidth]{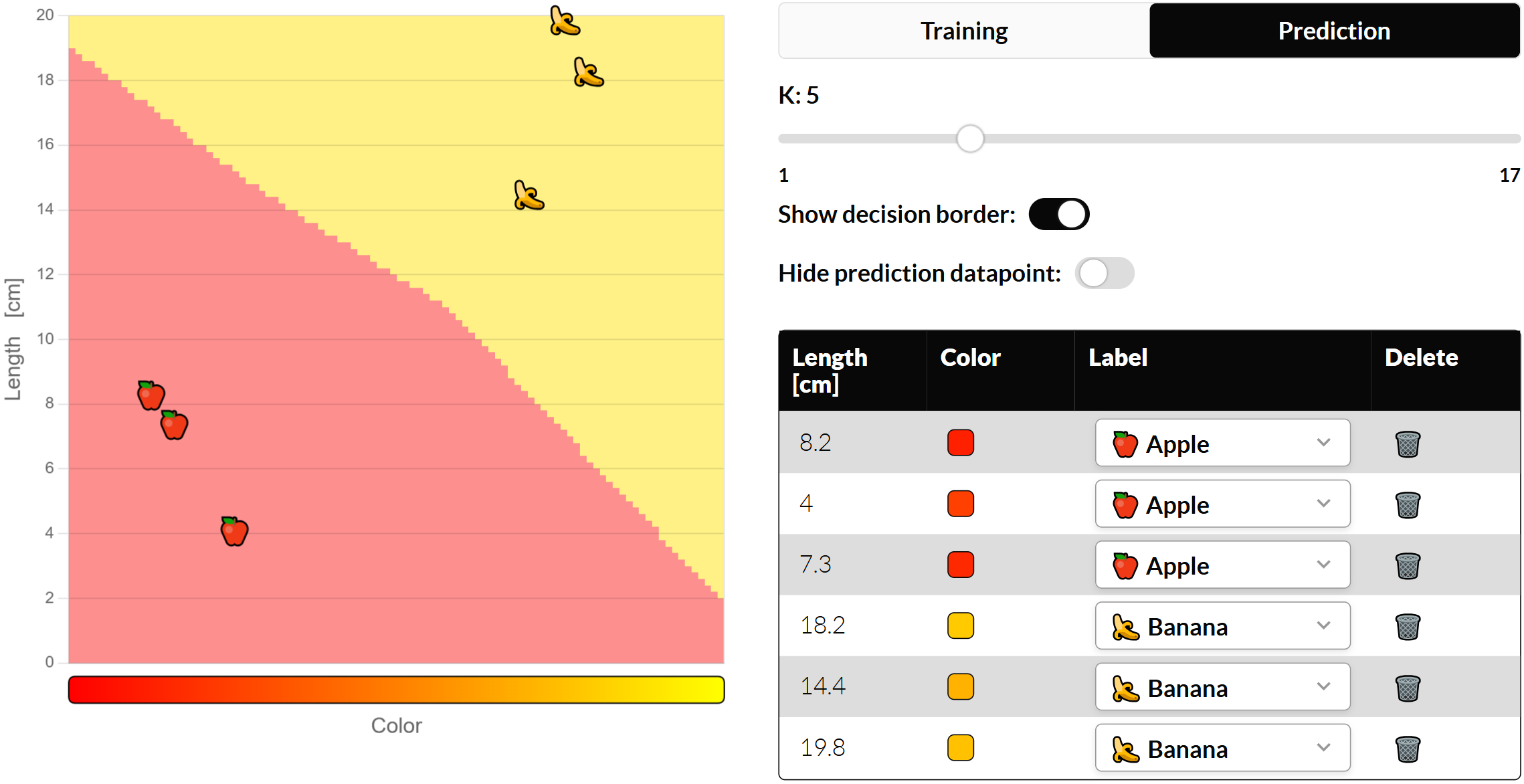}
    \caption{User interface of the \textit{Fruit Detector} experiment in inference mode. The 2D plot visualizes collected samples and the decision boundary. The interface also includes a data table for editing or deleting samples, and controls for switching between training and inference modes and adjusting the number of neighbors participating in the vote.}
    \label{fig:fruit_detector_ui}
\end{figure}

\textbf{Web interface.} The web interface visualizes collected data on a 2D plot with color on the x-axis and length on the y-axis as shown in \Cref{fig:fruit_detector_ui}. The interface operates in two modes: training and inference. In training mode, users select the fruit class and record color and length measurements. In inference mode, an unknown fruit can be classified, and connections to its nearest neighbors are displayed. Users can also adjust the number of nearest neighbors considered in the analysis, K in KNN, and toggle the decision boundary. A data table provides an overview of all collected samples, which allows users to edit labels or delete datapoints.

\textbf{Educational scaffolding.} Participants first brainstorm possible features to distinguish apples from bananas. After the discussion, color and length are introduced as the selected measurable features. The children then collect a small dataset by measuring at least three samples of each fruit. To build intuition, an exemplary plot is shown with existing data. A question-mark marker representing an unknown fruit is placed in different positions, and participants are asked to predict its class. This naturally leads to the idea of proximity in KNN, after which the algorithm is introduced.

The participants then test the model in inference mode, observe predictions, and explore its limitations, such as when measuring an orange that falls outside the trained classes. Finally, by intentionally poisoning the data, they see how the decision boundary becomes distorted, highlighting the importance of accurate labeling and data quality.

\subsection{Pitcher}

\begin{figure}[!ht]
    \centering
    \includegraphics[width=\columnwidth]{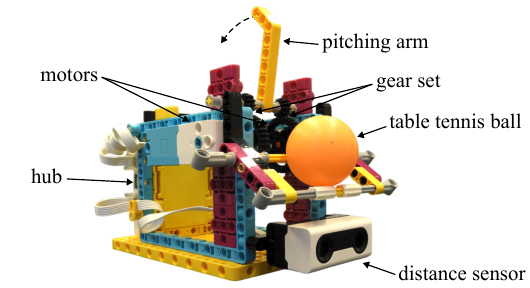}
    \caption{\textit{Pitcher} consisting of a hub, a distance sensor, and motors that, through a gear set, accelerate a pitching arm to launch a table tennis ball toward a target.}
    \label{fig:pitcher_photo}
\end{figure}

\textbf{Device design.} The \textit{Pitcher} launches a table tennis ball so that the ball hits a target placed in front of it. Its main components are displayed in \Cref{fig:pitcher_photo}. By varying motor speed, the flight distance can be controlled, and a linear mapping between motor speed and distance provides a reasonable approximation of this relationship.

\begin{figure}[!ht]
    \centering
    \includegraphics[width=\columnwidth]{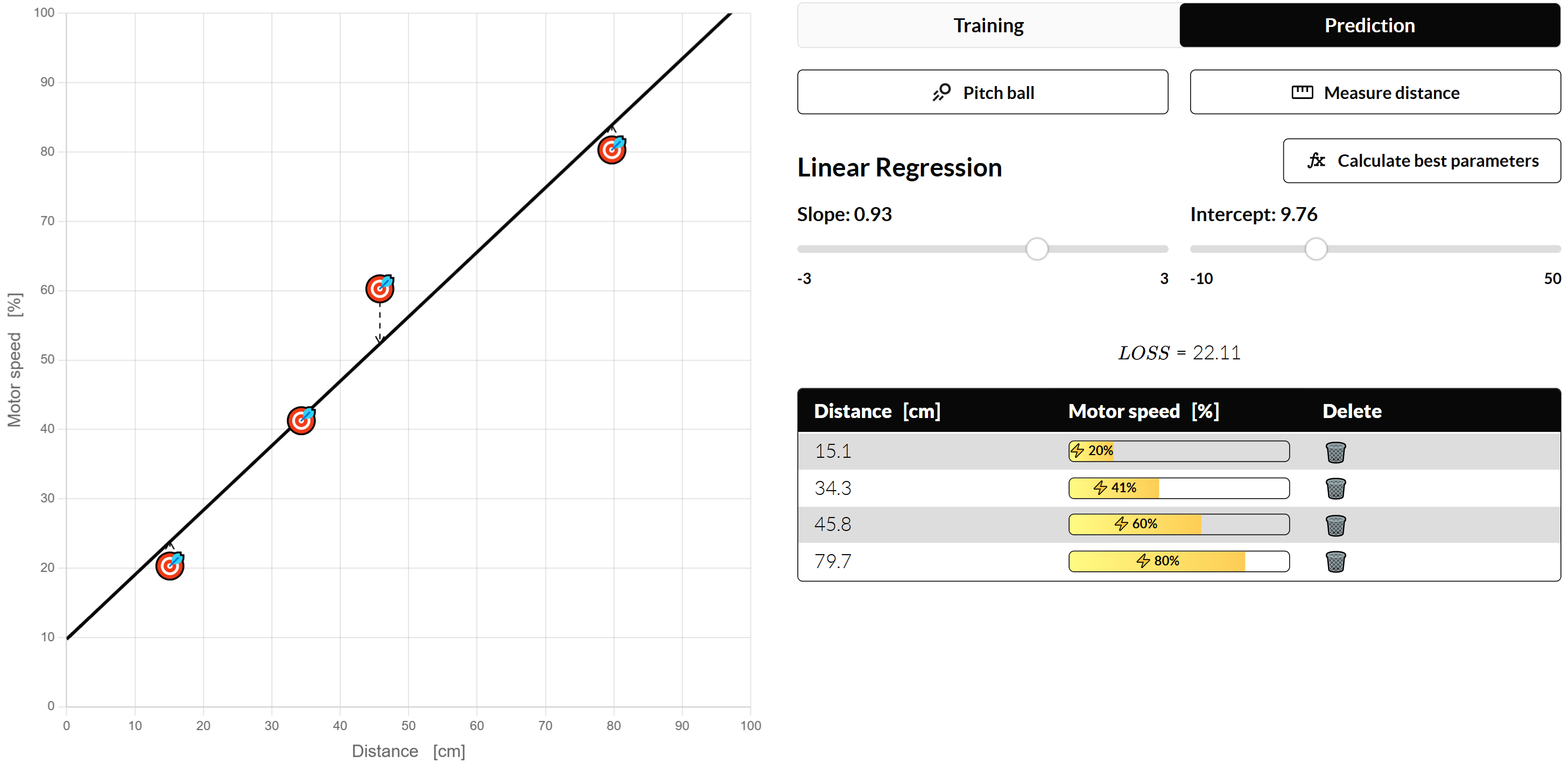}
    \caption{User interface of the \textit{Pitcher} experiment in inference mode. The 2D plot visualizes collected data points and the fitted regression line. The interface includes a data table for managing measurements and controls for adjusting line parameters, launching the ball, measuring distance, switching between training and inference modes, and calculating the best-fit line automatically.}
    \label{fig:pitcher_ui}
\end{figure}

\textbf{Web interface.} The interface is shown in \Cref{fig:pitcher_ui}. It displays a plot with motor speed on the y-axis and distance to the target on the x-axis, accompanied by a table of collected data points and buttons to launch the ball and measure the distance to the target or landing point. The system has training and inference modes: in training, users can adjust motor speed to record measurements at different values, while in inference, they can adjust the parameters of the line equation representing the speed–distance relationship, observe the resulting loss, and use a button to automatically calculate the best-fit solution.

\textbf{Educational scaffolding.} Children are first introduced to the core idea of the device. They launch the ball at different motor speeds and record the resulting flight distances, observing that the pattern forms a line. This motivates an introduction to linear regression and the loss function, which measures how closely the line fits the data. On the website, children modify the line parameters to minimize the loss, then let the computer automatically calculate the optimal line. Finally, they use the line to predict the motor speed required to hit a target with the ball.\looseness=-1

\subsection{Crawler}
\begin{figure}[!ht]
    \centering
    \includegraphics[width=\columnwidth]{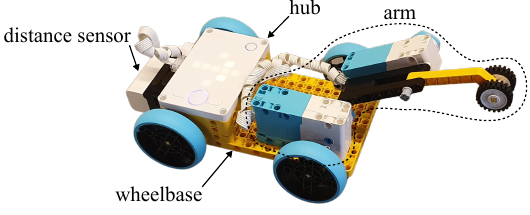}
    \caption{\textit{Crawler} consisting of a hub and a distance sensor mounted on a wheeled base, actuated by a two-limb arm that enables forward movement.}
    \label{fig:crawler_photo}
\end{figure}

\textbf{Device design.} The \textit{Crawler}, shown in \Cref{fig:crawler_photo}, is a wheeled platform equipped with a two-limb arm that enables forward movement through Q-learning. The arm is constrained to four discrete positions, with actions corresponding to transitions between these positions. A distance sensor measures the crawler’s position relative to a reference wall, allowing the calculation of distance traveled as a result of each action, which serves as the reward.

\begin{figure}[!ht]
    \centering
    \includegraphics[width=\columnwidth]{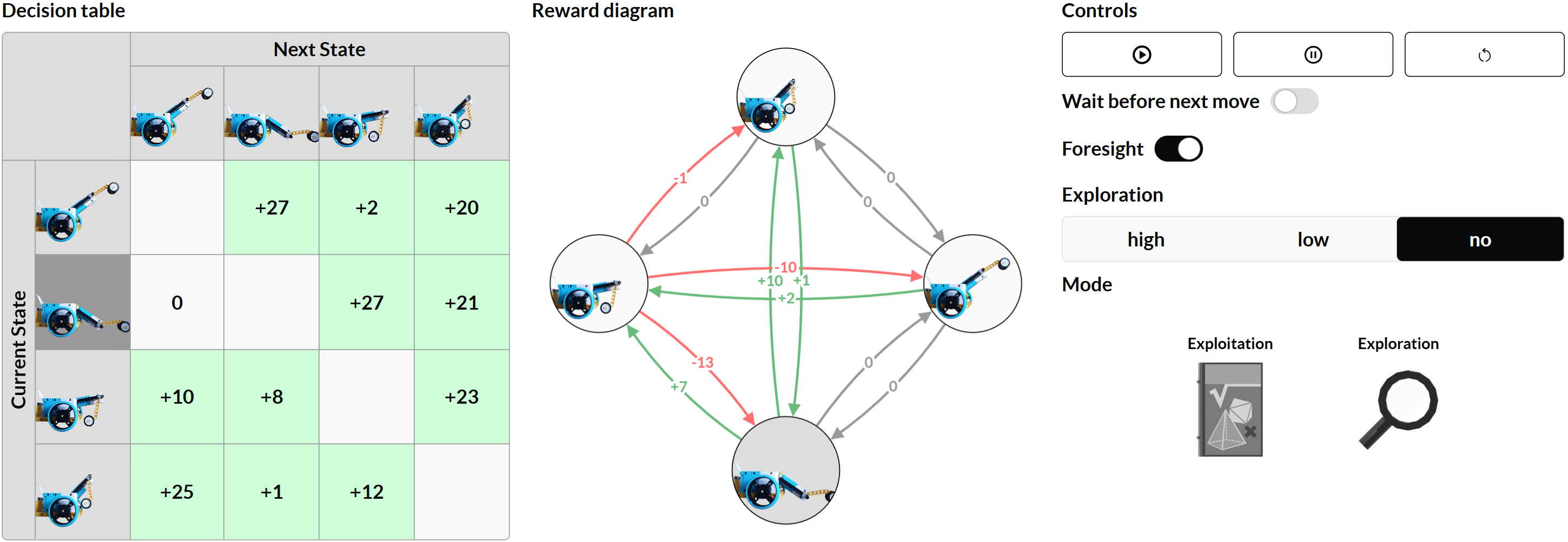}
    \caption{User interface of the \textit{Crawler} experiment displaying the Q-table (left), the reward diagram illustrating distance crawled for each transition (center), and the control panel (right), which includes options to pause, resume, or reset training, adjust the exploration rate, and toggle the discount factor to account for future rewards.}
    \label{fig:crawler_ui}
\end{figure}

\textbf{Web interface.} The interface is displayed in \Cref{fig:crawler_ui} and includes a Q-table, a reward diagram showing the distance traveled for each state transition, and a control panel. Users can pause, continue, or reset the experiment, adjust the exploration rate (with an indicator that shows whether the current action is exploratory or exploitative), and toggle the discount factor between 0 and 1 to account for future rewards.

\textbf{Educational scaffolding.} After a brief introduction to the fundamentals of reinforcement learning, children assemble the \textit{Crawler} and learn that the quality of each action depends on the distance the \textit{Crawler} moves, which is determined by the current state and the chosen action. This motivates the introduction of the Q-table, after which training of the robot~begins.

Initially, the \textit{Crawler} struggles to explore the state-action space, leading to the introduction of the exploration-exploitation dilemma. The \textit{Crawler} is trained with high exploration, and once sufficient experience is gathered, exploration is reduced to zero. At this stage, the \textit{Crawler} converges to a suboptimal solution, highlighting the importance of accounting for future rewards.

Finally, training is repeated with the discount factor enabled, allowing the \textit{Crawler} to consider future rewards. After gradually reducing exploration, the \textit{Crawler} successfully learns an optimal movement strategy, demonstrating how expected future returns improve performance.

\section{Course Design}
\label{sec:course_design}

This section presents the workshop built on the basis of our web-based platform to introduce core machine learning concepts to K–12 students. The following subsections describe the course’s target audience and learning objectives, pedagogical framework, session structure, and required tools and resources.

\subsection{Target Audience and Objectives}
This course targets students aged 12-17, an age range selected to ensure participants possess a basic understanding of coordinate systems and linear equations -- mathematical foundations essential for grasping the concepts introduced. The curriculum is designed to accommodate varying levels of prior experience in robotics and programming, including students with no prior exposure to these fields.

The objectives of this course include:
\begin{itemize}
    \item Understanding the core principles of machine learning including pattern recognition and algorithmic \mbox{decision-making}.
    \item Exploring fundamental machine learning algorithms such as KNN, linear regression, and Q-learning.
    \item Applying machine learning concepts to real-world scenarios through interactive robotics demonstrations.
    \item Developing problem-solving and data analysis skills.
    \item Achieving AI literacy by understanding strengths \mbox{and limitations}.
    \item Motivating continued exploration in machine learning.
\end{itemize}

\subsection{Pedagogical Framework}

The course is designed to be hands-on, grounded in the principles of constructionism \cite{papert1991, papert1980} and collaborative learning \cite{chi2014}, and incorporates multiple strategies to promote inclusivity, engagement, and effective learning.

To promote \textbf{inclusivity}, students collaborate in pairs with shared robotics kits, fostering peer interaction and cooperative problem-solving. Furthermore, minimal prerequisites in form of the aforementioned basic mathematical concepts eliminate experience gaps and ensure accessibility for students with diverse backgrounds.

To maximize both \textbf{engagement} and \textbf{learning effectiveness}, the curriculum employs several integrated strategies. An iterative theory-practice cycle alternates concise theoretical instruction with hands-on activities in which students immediately apply the concepts discussed. A no-code approach utilizing web-based interface allows students to focus on algorithmic understanding rather than syntax and implementation details. Post-session quizzes with incentive-based rewards reinforce learning while simultaneously motivating attentiveness during theoretical segments and enabling formative assessment.

\subsection{Course Structure}
The course consists of two four-hour sessions with refreshment breaks, conducted on separate days. In our experience, the full program, including breaks, can be completed in six to seven hours. Alongside brief theoretical input, the course includes three major hands-on sections that use the robots described earlier to visualize machine learning concepts and algorithms. The course is structured as follows:

\begin{itemize}
\item Day 1:
    \subitem \textbf{Introduction:}
    \begin{itemize}
        \item Course overview 
        \item Ice-breaker activity
    \end{itemize}
    \subitem \textbf{Machine learning fundamentals:}
    \begin{itemize}
        \item Machine learning as pattern recognition
        \item Machine learning's strengths over humans
        \item Core concepts: model, data, features, and labels
    \end{itemize}
    \subitem \textbf{\textit{Fruit Detector:}}
    \begin{itemize}
        \item Feature selection for fruit classification
        \item KNN and inference
        \item Edge cases unseen in the training set
        \item Decision boundary and data quality
    \end{itemize}
    \subitem \textbf{Quiz}
\item Day 2:
    \subitem \textbf{\textit{Pitcher:}}
    \begin{itemize}
        \item Classification vs. regression tasks
        \item Device assembly and data collection
        \item Linear regression and loss function
        \item Inference
    \end{itemize}
    
    \subitem \textbf{Machine learning paradigms:}
    \begin{itemize}
        \item Supervised vs. unsupervised learning
        \item Applications of unsupervised learning
        \item Reinforcement learning and its applications
    \end{itemize}
    
    \subitem \textbf{\textit{Crawler:}}
    \begin{itemize}
        \item Core reinforcement learning concepts: reward, environment, state, and action
        \item Robot assembly and training
        \item Exploration vs. exploitation trade-off
        \item Accounting for future reward in decision-making
    \end{itemize}
    \subitem \textbf{Conclusion:}
    \begin{itemize}
        \item Course recap and key takeaways
        \item Closing remarks
    \end{itemize}
    \subitem \textbf{Quiz}
\end{itemize}

\subsection{Tools and Resources}
Students work in pairs, each provided with a LEGO Education Spike Prime kit and a shared laptop. Each pair receives A3 sheets containing printed instructions from the website for device assembly, hub connection, and experiment procedures. In addition, students are supplied with all materials needed for the experiments: apples, bananas, and an orange for the \textit{Fruit Detector}, and table tennis balls and a distance measurement mat for the \textit{Pitcher}.

\section{Study Design}
\label{sec:study_design}

To evaluate the effectiveness of our web-based platform and workshop curriculum, we conducted an empirical study with K-12 students during two iterations of a community outreach workshop. This section describes our research methodology, including participant recruitment procedures, ethical protocols, measurement instruments, data collection procedures, and participant characteristics. \looseness=-1

\subsection{Participant Recruitment and Selection}
The course was conducted twice at a local makerspace during school holidays as a community outreach workshop for youth aged 12 to 17. Each iteration consisted of a two-day workshop with four sessions. Participation was advertised through the makerspace's media channels and limited by available space and the number of LEGO kits. Enrollment was allocated on a first-come, first-served basis. Prior to the course, families received information about the voluntary anonymous study along with consent forms. Out of 23 course participants, 14 brought signed consent forms and agreed to participate in the study.\looseness=-1

\subsection{Ethical Considerations}
All procedures followed the approved institutional ethics board protocol. Consent forms were provided to fully inform both parents and youth participants before the course. Course instructors explicitly explained the study procedures in person and addressed participant questions. Written informed consent was obtained from all participants prior to data collection, ensuring voluntary and informed participation. Participants were informed they could withdraw from the study at any point. Anonymity was maintained throughout the study by not collecting any identifying information from participants.

\subsection{Data Collection}
Pre- and post-course surveys were administered on paper, facilitated by two course instructors. Participants completed the pre-survey independently before the course began (approximately 10 minutes) and the post-survey independently immediately following the course conclusion (approximately 10 minutes). Instructors provided clarifications as needed but stepped away during survey completion to ensure privacy. All students in the class, regardless of study participation, remained together in the same room and engaged in identical learning activities. All participants who provided consent completed and submitted both surveys.

\subsection{Survey Instruments}
Pre- and post-course surveys were administered in German to assess participants' knowledge, perceptions, and course~experience.\looseness=-1

The \textbf{pre-course} survey collected demographic information (age, career interests in machine learning/robotics/AI, general interests, and club memberships) and assessed baseline knowledge. Participants listed three to five words associated with the term `AI' and rated their familiarity with eight machine learning concepts using a 5-point Likert scale \cite{likert1967} shown in~\Cref{tab:learning_outcomes_scale}.

\begin{table}[!ht]
\centering
\caption{5-point Likert scale used to assess participants' self-reported knowledge of machine learning concepts.}
\begin{tabular}{cl}
\toprule
\textbf{Score} & \textbf{Description} \\
\midrule
1 & Never heard of it \\
2 & Only heard the name \\
3 & Heard something about it, but didn't really understand \\
4 & Have a bit of intuition for it \\
5 & Can explain it well enough to teach it to a friend \\
\bottomrule
\end{tabular}
\label{tab:learning_outcomes_scale}
\end{table}

The \textbf{post-course} survey repeated all pre-course questions to measure changes in knowledge and AI-related terminology. Additional items evaluated course design, engagement, and impact using a standard 5-point Likert scale (1 = strongly disagree, 5 = strongly agree). An open-ended question collected additional feedback from participants.

\subsection{Participants}
The majority were middle-school students, with only a small number representing high school levels. In terms of personal interests, most participants expressed an affinity for technology and science, followed by art and culture, sports, and crafts. Nearly all participants were members of sports clubs.

\section{Results}
\label{sec:results}

We present findings from pre- and post-course surveys completed by 14 workshop participants. Results are organized into four subsections examining learning outcomes, AI-related terminology, course design effectiveness, and motivation for future AI engagement.

\subsection{Learning outcomes}

As shown in \Cref{tab:learning_outcomes}, participants' self-assessed knowledge levels improved across all topics. The largest gains were observed for questions concerning machine learning paradigms, their distinctions, and their applications (L2 and L3), both of which showed statistically significant improvements ($p < 0.01$). Similarly, self-assessed knowledge of specific algorithms and related concepts, including KNN, linear regression, reinforcement learning, and the exploration-exploitation dilemma (L5, L6, L7, L8), also increased, with all gains reaching statistical significance ($p < 0.05$). Topics with pre-existing higher familiarity, such as general machine learning concepts (L1) and the importance of data in machine learning (L4), also showed increases in mean scores, although these improvements did not reach conventional statistical significance. Overall, these results suggest that the course was associated with gains in understanding of specific machine learning paradigms and algorithms, while also showing positive trends for more general concepts.

\begin{table*}[!ht]
\centering
\caption{Self-assessed knowledge on a 5-point Likert scale from \Cref{tab:learning_outcomes_scale}. Sample sizes vary due to missing responses. Statistical significance determined using two-sided Mann-Whitney U tests. * $p \leq 0.05$, ** $p \leq 0.01$}
\small
\begin{tabular}{lccccc}
 \toprule
 \textbf{Concept} & $n_{\text{pre}}$ & $\text{M}_\text{pre}$ (SD) & $n_{\text{post}}$ & $\text{M}_\text{post}$ (SD) & $p$-value \\
\midrule
L1. Machine learning & 14 & 3.14 (1.03) & 14 & 3.86 (1.17)  & 0.066 \\
L2. Three machine learning paradigms and their differences** & 14 & 1.79 (1.05) & 14 & 3.21 (0.97)   & 0.002 \\
L3. Where each of the machine learning paradigms can be applied** & 14 & 1.64 (0.93) & 14 & 3.07 (1.27)  & 0.003 \\
L4. Data and its importance in machine learning & 13 & 2.46 (1.27) & 13 & 3.54 (1.45)   & 0.063 \\
L5. KNN** & 12 & 1.25 (0.87) & 13 & 3.23 (1.54)   & 0.001 \\
L6. Linear regression** & 13 & 1.23 (0.60) & 13 & 2.92 (1.50)   & 0.003 \\
L7. Reinforcement learning* & 13 & 1.85 (1.14) & 14 & 3.14 (1.23)  & 0.014 \\
L8. Exploration-exploitation dilemma* & 13 & 2.08 (1.19) & 13 & 3.23 (1.48)  & 0.043 \\
 \bottomrule
\end{tabular}
\label{tab:learning_outcomes}
\end{table*}

\subsection{AI-Related Terminology}

\Cref{tab:ai_perception} illustrates changes in participants' AI-related terminology between the pre- and post-assessment. While general domain-related terms, as well as descriptions of applications, perceptions, and abstract concepts, remained largely consistent, the word ``danger" disappeared from post-assessment responses. At the same time, participants began using specific machine learning terms, including ``agent," ``rewards," ``data," ``linear regression," and ``supervised learning", which suggests increased familiarity with algorithm-specific vocabulary.

\begin{table}[!ht]
\centering
\caption{Changes in participants' AI-related terminology between pre- and post-survey responses. Words and phrases are categorized thematically based on content analysis of open-ended responses asking participants to list 3-5 words associated with the term `AI'. Two pre-survey responses yielded unclear entries and were removed.}
\begin{tabular}{p{0.15\columnwidth} p{0.35\columnwidth} p{0.35\columnwidth}}
\toprule
&\textbf{Pre-survey} & \textbf{Post-survey} \\
 \midrule

\textbf{Domain terms} & artificial intelligence, machine learning, artificial & AI, artificial intelligence, artificial, machine learning \\
 \midrule
\textbf{Abstract concepts} &future, science, technology, learning, progress, development, try out & progress, science, future, technology, tech, learning \\
 \midrule
\textbf{Perception} & intelligent, efficiency, potential danger, clever & intelligent, child \\
 \midrule
\textbf{Application} &autonomous driving, image analysis, image recognition, photo editing, helper, help with research, search for information, translate, ChatGPT, robot, robotics, programming, machines, JavaScript, assistant, independent learning & image editing, images, ChatGPT, program a code, help, help with working, information research, programming, robot, smart assistant, support for machines \\
 \midrule
\textbf{Machine learning terms} & -- & agent, rewards, data, action, linear regression, models, environment, supervised learning \\
 \bottomrule
\end{tabular}
\label{tab:ai_perception}

\end{table}

\subsection{Course design}

As shown in \Cref{tab:course_design}, participants rated the course highly across all evaluated aspects, with mean scores ranging from 4.00 to 4.54 on a 5-point Likert scale. In particular, the clarity and understandability of the presentations and examples (D5) and the enjoyment of hands-on activities (D7) received the highest ratings (both 4.54), suggesting strong engagement and positive reception of the instruction. Other aspects, including the mathematical content (D1), the clarity and interest of data visualizations (D2-D3), and platform usability (D4), were also rated positively, suggesting that the course was both accessible and well-structured.

\begin{table}[!ht]
\centering
\caption{Course evaluation ratings on a standard 5-point Likert scale. Sample sizes vary due to missing responses.}
\small
\begin{tabular}{p{6cm} cc}
\toprule
\textbf{Question} & $n$ & M (SD) \\
\midrule
D1. I was able to follow the mathematics in the course & 13 & 4.08 (0.64) \\
D2. The data visualizations were clear & 12 & 4.08 (0.79) \\
D3. The data visualizations were interesting & 13 & 4.38 (0.77) \\
D4. The web platform is intuitive and easy to use & 12 & 4.00 (1.04) \\
D5. The presentations and examples were easy to understand & 13 & 4.54 (0.66) \\
D6. I would recommend this course to my friends & 13 & 4.15 (0.69) \\
D7. I found the devices we built fun and interesting & 13 & 4.54 (0.52) \\
\bottomrule
\end{tabular}
\label{tab:course_design}
\end{table}

Participants also reported learning new AI concepts, enjoying the LEGO-based hands-on activities, and finding the course both fun and worth recommending to peers as a response to the open-ended feedback question.

\subsection{Motivation for AI}

Participants reported high motivation for further engagement with AI and machine learning. As shown in \Cref{tab:motivation_ai}, the desire to continue exploring machine learning after the course (E1) received a mean score of 3.92, while participants rated the usefulness of skills learned for future studies or careers (E2) at 3.69.

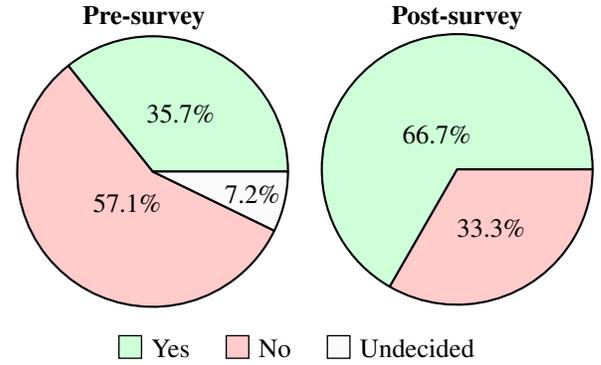
\begin{figure}[t]
\centering

\input{figures/tikz/career_interest_change.tikz}

\caption{Shift in participants' interest in pursuing a career in AI, machine learning, or robotics from pre-survey $(n = 14)$ to post-survey $(n = 12)$.}
\label{fig:career_interest_change}
\end{figure}

In addition, a higher number of participants expressed interest in pursuing a career in AI. As shown in \Cref{fig:career_interest_change}, the number of participants expressing interest in an AI/machine learning/robotics career rose from 5 pre-course to 8 post-course. Similarly, participants’ thematic interests shifted: in the post-survey, all participants indicated an interest in technology and science, compared to 11 out of 14 in the pre-survey.

\begin{table}[!ht]
\centering
\caption{Motivation for continued AI engagement on a standard 5-point Likert scale. Sample sizes vary due to missing~responses.\looseness=-1}
\small
\begin{tabular}{p{6cm} c c}
\toprule
\textbf{Question} & $n$ & M (SD) \\
\midrule
E1. I want to continue exploring machine learning after the course & 13 & 3.92 (1.04) \\
E2. The skills I learned in this course will be useful in my future studies or career. & 13 & 3.69 (0.63) \\
\bottomrule
\end{tabular}
\label{tab:motivation_ai}
\end{table}

\section{Discussion}
\label{sec:discussion}

This study evaluated the effectiveness of a web-based platform and two-day workshop curriculum to teach machine learning concepts to K-12 students through programming-free robotics activities. Our findings suggest that combining interactive visualizations with tangible LEGO robotics can make basic machine learning algorithms accessible and engaging for young learners. This section interprets our key findings and acknowledges the limitations of the study.

\subsection{Effectiveness of Visualization-Based Robotics Approach}
Statistically significant improvements in students' self-assessed understanding of machine learning paradigms and specific algorithms (L2, L3, L5-L8) are consistent with our hypothesis that tangible, visualization-based approaches can effectively teach technical machine learning concepts to young learners. The particularly strong gains in understanding machine learning paradigms and their applications suggest that the hands-on, multi-algorithm approach helped students develop a broader conceptual framework for machine learning, rather than isolated knowledge of individual techniques.

The non-significant gains in general machine learning knowledge (L1) and data importance (L4) likely reflect participants' prior exposure to these foundational concepts combined with the study's limited sample size.

\subsection{Shift to Technical Understanding}
The emergence of algorithm-specific terminology in post-course responses (agent, rewards, data, linear regression, supervised learning) suggests increased familiarity with core machine learning concepts. The shift from general descriptors toward more technical language is consistent with research showing that revealing the inner workings of machine learning models can improve understanding \cite{kulesza2015}, though more rigorous assessment would be needed to confirm deeper conceptual~change.\looseness=-1

\subsection{Platform Design and Usability}

The high scores for platform usability (D4: M = 4.00), visualization clarity (D2: M = 4.08), and instructional comprehensibility (D5: M = 4.54) support our design decisions to eliminate programming requirements and prioritize intuitive interfaces. Consistently positive feedback on hands-on activities (D7: M = 4.54) suggests that LEGO robotics served as an engaging medium for abstract concepts, supporting the constructionist approach \cite{papert1991, papert1980}.

\subsection{Motivation and Career Interest}

The increase in AI career interest (from 5 to 8 participants) and universal post-course interest in technology and science suggest the workshop was associated with increased curiosity about machine learning and related fields. The high motivation scores (E1: M = 3.92, E2: M = 3.69) indicate that participants perceived both the intrinsic interest and practical value of machine learning skills, which may support sustained engagement beyond the workshop.

\subsection{Limitations}

Several limitations warrant consideration when interpreting these findings:

\begin{itemize}
    \item \textbf{Sample size and generalizability.} With only 14 participants from a single location, our results may not generalize to broader populations. The self-selected nature of workshop attendance may have attracted students with pre-existing interest in technology, potentially \mbox{limiting representativeness}.
        
    \item \textbf{Self-reported measures.} Our reliance on self-assessed knowledge may not fully capture actual conceptual understanding. Future research should incorporate objective assessments such as performance tasks or transfer tests.
    
    \item \textbf{Short-term evaluation.} We assessed outcomes immediately post-course, precluding conclusions about knowledge retention or long-term motivation. Follow-up studies should examine whether gains persist over time.
    
    \item \textbf{Varying sample sizes.} Missing responses across different items ($n = 12-14$) reduce statistical power and may introduce bias.
    
    \item \textbf{Lack of control group.} Without a comparison condition, we cannot definitively attribute learning gains to our specific approach.
\end{itemize}

Despite these limitations, our findings provide initial evidence that the \textit{Machine Learning with Bricks} platform and curriculum can serve as an engaging introduction to machine learning concepts for young learners, with promising indicators of motivation for continued learning.

\section{Conclusion}
\label{sec:conclusion}

This paper presents \textit{Machine Learning with Bricks}, an open-source web-based platform that combines interactive visualizations with LEGO robotics to teach core machine learning algorithms to K-12 students without requiring programming skills. Through a two-day workshop with 14 students aged 12-17, we provide initial evidence that this approach can make basic machine learning concepts, such as KNN, linear regression, and Q-learning, accessible and engaging for young learners.\looseness=-1

Our empirical evaluation revealed four key outcomes. First, participants achieved statistically significant improvements in self-assessed understanding of machine learning paradigms and specific algorithms, with particularly strong gains in areas that were novel to them. Second, students' AI-related vocabulary shifted toward more algorithm-specific terminology. Third, the platform received high usability ratings and the hands-on activities were consistently engaging. Fourth, the workshop was associated with higher reported motivation for continued AI learning and interest in pursuing AI-related careers.\looseness=-1

These findings contribute to the growing body of research on AI education. By eliminating programming requirements and leveraging tangible robotics with real-time visualization, we provide initial evidence that young learners can grasp machine learning concepts and algorithms while maintaining high engagement and motivation. 

\subsection{Future Work}

Several promising directions emerge from this work:

\begin{itemize}
    \item \textbf{Curriculum expansion.} Future versions could incorporate additional algorithms such as boosting, neural networks, and decision trees, allowing students to compare performance across different models. Unsupervised learning could be introduced through clustering activities using fruits as data points.
    
    \item \textbf{Competency development.} The curriculum could be extended to address a broader range of AI literacy competencies \cite{long2020}, including deeper treatment of ethics, fairness, and societal implications alongside the technical~content.\looseness=-1
    
    \item \textbf{Scale and longitudinal evaluation.} The platform is being integrated into the Robotics in Germany (RIG) cluster education track, which will enable larger-scale studies, objective knowledge assessments, and follow-up evaluations to provide stronger evidence of learning effectiveness and knowledge retention over time.
    
\end{itemize}

\section*{Acknowledgements}

The authors thank the KI-Makerspace T\"ubingen for hosting the workshops and all the students who participated in the study. We gratefully acknowledge Yufeng Wu for answering our questions regarding the Bluetooth connection to the LEGO SPIKE Prime hub. The robotic designs were inspired by the work of Prof. Val Rousseau (Crawler), Rebecca Shen (Pitcher), and the Tufts Center for Engineering Education and Outreach (Fruit Detector). Michael Muehlebach thanks the German Research Foundation and the Branco Weiss Fellowship, administered by ETH Zurich, for their financial support.

\bibliographystyle{IEEEtranN}
\bibliography{references}

\end{document}

%% file: figures/tikz/career_interest_change.tikz
\newcommand{\colorboxlabel}[2]{%
    \tikz[baseline=-0.5ex]\draw[fill=#1, draw=black, line width=0.5pt] (-.1cm,-.1cm) rectangle (0.2cm, 0.2cm); #2%
}

\begin{tabular}{cc}
    \textbf{Pre-survey} & \textbf{Post-survey} \\
     \begin{tikzpicture}
         \pie[color={light_green, light_red, light_gray}, 
              text=, 
              radius=1.8, 
              sum=100]
              {35.7/Yes, 57.1/No, 7.2/Undecided}
     \end{tikzpicture}
         &
     \begin{tikzpicture}
         \pie[color={light_green, light_red, light_gray}, 
              text=, 
              radius=1.8, 
              sum=100]
              {66.7/Yes, 33.3/No} 
     \end{tikzpicture} \\
     \multicolumn{2}{c}{
        \colorboxlabel{light_green}{Yes} \quad
        \colorboxlabel{light_red}{No} \quad
        \colorboxlabel{light_gray}{Undecided}
    } 
\end{tabular}